\documentclass{article}
\usepackage{spconf,amsmath,graphicx}

\usepackage{times}
\usepackage{latexsym}

\usepackage{url}

\usepackage{amsmath}
\usepackage{graphicx}
\usepackage{multirow}
\usepackage{color}

\usepackage{algorithmicx}
\usepackage[ruled]{algorithm2e}
\usepackage{algpseudocode}
\usepackage{makecell}
\usepackage{multirow}
\usepackage{diagbox}

\title{Suffix Retrieval-Augmented Language Modeling}
\name{Zecheng Wang and Yik-Cheung Tam}
\address{NYU Shanghai\\
  Department of Computer Science \\
  567 West Yangsi Road, Pudong New District, Shanghai 200126, China
}

\begin{document}
\ninept
\maketitle

\begin{abstract}
Causal language modeling (LM) uses word history to predict the next word. BERT, on the other hand, makes use of bi-directional word information in a sentence to predict words at masked positions. While BERT is effective in sequence encoding, it is non-causal by nature and is not designed for sequence generation. In this paper, we propose a novel language model, {\bf SU}ffix {\bf RE}trieval-{\bf A}ugmented {\bf LM} (SUREALM), that simulates a bi-directional contextual effect in an autoregressive manner. SUREALM employs an embedding retriever to search for training sentences in a data store that share similar word history during sequence generation. In particular, the suffix portions of the retrieved sentences mimick the ``future'' context. We evaluated our proposed model on the DSTC9 spoken dialogue corpus and showed promising word perplexity reduction on the validation and test set compared to competitive baselines. Our source code is released on GitHub~\footnote{\url{https://github.com/Victor-wang-902/SUREALM.git}}.

%We also shed some light on dynamic LM adaptation to new domain when additional text from the new domain are added into the embedding retriever without language model retraining.
\end{abstract}
\section{Introduction}
\label{sec:intro}
Causal language modeling possesses great flexibility among most natural language tasks due to its unsupervised and generative nature. Large-scale pre-training on Transformer architecture like GPT2 has resulted in powerful models capable of capturing general knowledge of natural language. However, unlike bidirectional language models such as BERT, RoBERTa, causal language modeling 
can only look at the word history to predict the next word. While this is mathematically sound, 
such left-hand contextual information may potentially hinder the language model from capturing semantic knowledge at its fullest. 
On the other hand, while BERT provides satisfactory performance for sequence encoding thanks to its bi-directional nature, it is designed for masked language modeling which predicts the word identity at a masked position in a sentence. BERT is non-causal and thus is not suitable for sequence generation.

% talk about motivation and our approach briefly
Recent studies have shown that retrieving prefix contextual information from an external data store can further improve performance of a causal language model without increasing the number of model parameters~\cite{realm}. However, the retrieved information are still uni-directional. In this paper, we propose a novel language model, {\bf SU}ffix {\bf RE}trieval-{\bf A}ugmented {\bf LM} (SUREALM), that employs an embedding retriever for suffix retrieval from a data store. During sequence generation, the current word history, referred as prefix in the rest of the paper, is submitted to an embedding retriever to search for similar prefixes in a data store. Then the corresponding suffixes of these training sentences are viewed as ``future'' context to guide sequence generation. The intuition is that sentences sharing a similar given prefix may probably have strong correlation on their suffixes. For example, ``how may i'' and ``how can i'' are similar prefixes. If the model also knows the complete reference sentence ``how can i help you'', then the model would tend to predict ``help you'' given a novel prefix ``how may i''. To exploit this assumption, we perform all possible splitting of each training sentence into a triple: a prefix, a word, and a suffix. We employ pre-trained sentence transformers~\cite{reimers-2019-sentence-bert} to encode the prefix and suffix of each training sentence to create an embedding data store. Then an embedding retriever such as FAISS~\cite{johnson2019billion} is employed for prefix-suffix embedding retrieval given an encoded prefix. The retrieved prefix-suffix embeddings are augmented into the word embedding inputs during sequence generation, achieving the causal language modeling with a simulated bi-directional effect. SUREALM is causal because it only uses word history for predicting the next word. SUREALM is simulated bi-directional because it exploits ``future'' context from other similar sentences.
%We then speculate that even for the state-of-the-art mass pre-trained language models, the generalization power 
%While the issue of unidirectional context in autoregressive language modeling still has not been addressed, we argue that 
%Obviously, we cannot provide the exact suffix for the model to attend to as this would defeat the purpose of achieving autoregressiveness. 

% talk about contributions
Our contributions are two-folded: First, we propose SUREALM, a new causal language model enhanced by prefix-suffix embedding retrieval to simulate a bi-directional effect for sequence generation.
Second, we perform extensive experiments and show effectiveness of our model on the DSTC9 dialogue corpus. 
%In particular, we show that pre-trained BERT-based sentence transformers checkpoints for prefix and suffix encoding can be ``mixed'' well with the backbone transformer language model initialized with GPT2 or BERT checkpoints after LM fine-tuning.
%Third, we shed some light on the possibility of dynamic language model adaptation on new domain without model retraining via updating the data store with additional in-domain prefix-suffix embeddings.
\vspace{-1.5mm}
\section{Related Work}
\label{sec:related_work}

Improving language model using retrieval technique is not new. ~\cite{milind99irlm} employs document retrieval to retrieve relevant documents which are used to create an adaptive language model and interpolating it with the background statistical N-gram language model. ~\cite{eck-etal-2004-language} employs information retrieval to perform language model adaptation for statistical machine translation. Once the language models are adapted, they are kept fixed during sequence generation.

Memory Networks (MemNNs~\cite{a27da5feb471466cb024242bf91426d3}) are a family of neural networks integrating a memory component which can be updated during training. To avoid large storage of external memory, MemNNs can be trained end-to-end~\cite{sukhbaatar2015end} by computing a compatibility score between a query and memory items , similar to an attention mechanism. SUREALM resembles MemNNs as we utilize an external data store as ``memory''. However, SUREALM's ``memory'' contains precomputed embeddings and is retrieved with an embedding retriever for computational efficiency. Like E2E MemNNs~\cite{sukhbaatar2015end}, we also distinguish between memory and query. We further construct our memory as prefix-suffix pairs enabling a more effective retrieval scheme by matching input query with prefixes.

Most recent development in language modeling is based on transformers~\cite{NIPS2017_3f5ee243}. BERT-based Masked language modeling~\cite{devlin-etal-2019-bert, DBLP:journals/corr/abs-1907-11692} exploits bi-directional information of a sentence to predict the word identity of the masked tokens. While BERT is effective in encoding sequences, it is not suitable for sequence generation due to its non-causal nature. Causal language modeling such as GPT2~\cite{Radford2018ImprovingLU} is uni-directional. Our proposed model attempts to retain the best of the two worlds as autoregressive and simulated bi-directional via augmentation of suffix embeddings during sequence generation.

One noticeable work for language modeling using embedding retrieval is nearest neighbor language model (KNN-LM)~\cite{khandelwal20generalization}. Their approach store dynamic information in an external knowledge base. During sequence generation, KNN-LM uses the current prefix to retrieve similar prefixes in the data store using embedding retrieval. The output probability distribution is estimated by looking at the corresponding next words in the retrieved prefixes. Such word probability distribution is linearly interpolated with the output word distribution from the causal transformer LM. While it has shown effectiveness in reducing word perplexity, their approach is uni-directional in terms of utilization of information for word prediction. Our proposed model enjoys the simulated bi-directional effect of utilizing ``future'' contextual information to guide sequence generation.

Another work is retrieval-augmented generation for question and answering~\cite{rag}. Their approach employs an embedding retrieval over the encoded document embeddings. Then the top-k retrieved document embeddings are viewed as latent variables for answer generation. These latent variables are marginalized in the generator within a sequence-to-sequence generation framework. Related work of using retrieval technique for language modeling pre-training and question answering also includes~\cite{realm}. Our proposed model differs from their approach that we do not employ marginalization on the top-k retrieved results. In contrast, our model counts on the attention mechanism to attend to all previously retrieved suffix embeddings such that the cross-entropy loss is minimized.

%~\cite{realm} employs an embedding retriever for language model pretraining followed by question answering via model fine-tuning.

% talk about neural cache LM (Apple and related paper)
\vspace{-3mm}
\section{Proposed approach}
\vspace{-2mm}

\label{sec:approach}
Our proposed approach extends causal language models with suffix retrieval. Denote a sentence $W=w_1w_2...w_N$. Then our model defines the negative log likelihood of $W$ as follows:
\begin{equation}
\label{eqn:surealm}
%\begin{split}
\resizebox{.91\hsize}{!}{$\mathcal{L}(W;\Theta) = -\log P(W;\Theta) = -\sum_{i=1}^N \log P(w_i|p_i,f(p_i;\Phi);\Theta)$}
%\end{split}
\end{equation}
where $p_i$ denotes the word history (or prefix) of the word token $w_i$. $f(p_i;\Phi)$ denotes a retrieval function parameterized by $\Phi$, to search for sentences that have similar prefixes in a data store. Then the suffixes of the retrieved sentences are augmented  into the language model via suffix embedding. Although the true future context is unseen in causal language models, we hypothesize that such future context may be estimated by leveraging sentences that share similar prefixes. Thus, our model, {\bf SU}ffix {\bf RE}trieval-{\bf A}ugmented {\bf LM} (SUREALM), achieves a bi-directional modeling as in BERT and still be able to generate sentences in an autoregressive manner as in GPT.
In summary, our proposed approach has three steps: (1) Data preprocessing and indexing; (2) SUREALM training; (3) SUREALM decoding. We describe the steps in Section~\ref{subsec:preprocess}--~\ref{subsec:decode}.
%with overall system workflow in Figure~\ref{fig:workflow}.

% commented by Wilson (Most contents absorbed into new text)
%Our proposed approach enables bi-directional training of autoregressive LMs by leveraging aggregated %history and future contexts given observed prefixes. During data pre-processing, we introduce the %indexing operation; during training, we introduce retrieval-based decoding. Inference is trivial given %our training algorithm. Indexing, retrieving, and decoding can be summarized as follows: %\begin{enumerate}
%    \item During indexing, a datastore is created consisting of all pairs of prefix-suffix %representations in the training data with a pre-defined incremental step. For each pair, the prefix %representation is indexed for similarity search. 
%    \item Unlike traditional training of a causal LM which involves a self-attention decoder only, we %retrieve prefix-suffix embedding pairs from the datastore corresponding to the current prefixes by %similarity search which are then used in cross-attention blocks.
%    \item During forward pass of the decoder, given an input query, we get cross-attention key and %value by appending retrieved prefixes and suffixes to the input embedding respectively. Then, %cross-attention heads will have access to: 1) history contexts from prefix representations and 2) %aggregated future contexts from retrieved suffix representations.
%\end{enumerate} 
%Details of our workflow are described in the following sections and pseudo code for training and %inference can be found in \ref{alg}.
\vspace{-2mm}

\subsection{Data pre-processing and indexing}

\label{subsec:preprocess}
Given a training corpus $\mathcal{D} = \{W\}$ containing a set of unique sentences $W$, each sentence $W$ generates all possible partitions of $W$ into 3 parts: prefix, current word, and suffix, denoted as $(p_i, w_{i}, s_i)$ where $p_i = w_1w_2...w_{i-1}$, $s_j = w_{i+1}...w_N$ with valid word position $0 < i < N $.
Motivated from masked language modeling, we exclude the current word $w_{i}$ so that 
each data entry for indexing into a data store is
a prefix-suffix pair $(p_i, s_i)$. This formulation enforces our model to use information from the prefix (left context) and the retrieved suffixes (``right context'' from other training sentences).
Thanks to the recent development in sentence embedding retrieval, we employ pre-trained sentence transformer~\cite{reimers-2019-sentence-bert} to encode prefixes and suffixes such that retrieval is based on similarity of prefix embeddings. Then the data store returns the prefix and suffix embeddings. The rationale of encoding variable-length prefix and suffix into a fixed dimensional vector is to make the prefix and suffix representation smoother to mitigate word-level noises that are irrelevant to the current prefix under consideration. Intuitively, $(emb(p_i), emb(s_i)) \in \mathcal{R}^{d}$ can be viewed as a key-value pair where $d$ is the embedding dimension. To preserve positional information in representation, we use absolute positions in the original sentence when computing the prefix and suffix embeddings. We employ FAISS~\cite{johnson2019billion} for embedding search.

The final number of prefix-suffix pairs to index is $O(M\cdot N)$ where $M$ is the number of unique training sentences and $N$ is the maximum sequence length of a sentence. Essentially, our model requires to perform embedding retrieval at every word position. Therefore, we introduce a hyperparameter $\delta$ to control the frequency of embedding retrieval. For example, if embedding retrieval occurs at time $t'$, then the next time to retrieve will be at time $t'+\delta$. This implies that during time interval $[t'+1,t'+\delta-1]$, all the previously retrieved suffix embeddings are reused to save computation. This allows us to explore the tradeoff between computation and accuracy.
Regarding the suffix representation, we apply suffix truncation assuming that word tokens in a suffix that are closer to prediction time $t$ may be more helpful. We introduce a hyper-parameter $m$ for suffix truncation so that a truncated suffix $s_i'=w_{i+1}w_{i+2}...w_{i+m}$ with $m < N$ is fed into a sentence transformer for encoding. When the number of tokens $N$ is large, we conjecture that a right $m$ may avoid an overly smoothed suffix embedding representation due to the pooling mechanism in a sentence transformer. Table~\ref{tbl:faiss} shows sample retrieval results using some prefixes as input queries.

% commented by Wilson (content absorbed)
%The resulting mechanism resembles Masked LM, as our LM tries to utilize bidirectional information to %fill in the "blank". We also truncates the suffix by $m$ as we speculate that information too far %ahead will not help the LM decide the next token. In addition, incremental step $N$ determines the %granularity of $\mathcal{DS}$. With a smaller $N$, vectors encoded from more variety of string lengths %will be stored. When projecting a suffix using $T_{\theta}$, we also preserve its positional %information in the original sentence such that the suffix embedding will also contain such %information. Table~\ref{str slc} shows an example of string slicing mentioned above. The index %operation is also illustrated in \ref{workflow}.

% Victor TODO
\begin{table}[htb]
\vspace{-4mm}
\centering
\caption{Samples of retrieval results using FAISS.}
%\resizebox{\textwidth}{15mm}{
\begin{tabular}{|c|c|c|}
\hline
{\bf Input Query} & {\bf Retrieved Word} & {\bf Retrieved Suffix} \\ 
\hline

'i also want'&'free'& 'wifi' \\
\hline
'i'd like to' & 'book' & 'this hotel' \\
\hline
'is the hotel  &&\\ equipped with an' & 'elevator' & 'for convenience?'\\
\hline
'i have several' & 'options' &'for you. would you \\&& like a particular area...' \\

\hline
\end{tabular}%}
\label{tbl:faiss}
\vspace{-4mm}
\end{table}
\vspace{-2mm}
\subsection{SUREALM training}
\label{subsec:training}

\subsubsection{Offline retrieval}
One complication in SUREALM is the embedding retrieval required for each prefix $p_i$ at each word position $i$. First, it would be computationally expensive to perform on-the-fly embedding retrieval during training. Since we freeze the sentence transformer for encoding, top-K similar prefix and suffix embeddings can be precomputed offline using FAISS to speed up training. $K$ is a hyper-parameter to determine the number of retrieved suffix embeddings to be included for SUREALM training. To avoid cheating, we exclude all the embedding results belonging to the same training sentence ID. Empirically, we found this step to be crucial for SUREALM to learn from additional suffix information from other similar training sentences.

%Given a prefix string $\text{pre}^*$, we retrieve top-$k$ embedding pairs from $\mathcal{DS}$ to be %used during %forward pass of the decoder.\begin{equation} \label{c}
%    \mathcal{C}_k(\text{pre}^*, T_{\theta}, \mathcal{I}, \mathcal{DS}) = \{(p_1,s_1), (p_2,s_2), \dots, %(p_k,s_k)\}
%\end{equation}  $\text{pre}^*$ is first encoded by $T_{\theta}$ and then passed into $\mathcal{I}$ for searching.

% commented by Wilson (content absorbed)
%Subsequently, we retrieve top-$k$ result pairs from $\mathcal{DS}$. $k$ is a hyper-parameter %that determines the %amount of context information given to the decoder. During training, retrieving for %a train prefix %$\text{pre}_{\text{train}}$ will result in exact matches: pairs from the exact same %source sentence would be %retrieved, which is not desirable. So, we ensure that the model learns %generality by excluding the embedding %pairs that come from the same source sentence. For simplicity, we %omit $T_{\theta}, \mathcal{I}, \mathcal{DS}$ in %\ref{c} and write it as $\mathcal{C}_k(\text{pre}^*)$.
\vspace{-2mm}
\subsubsection{Mini-batch training}
\begin{figure}[htb]\label{fig:workflow}
\begin{minipage}[b]{\linewidth}
  \centering
  \centerline{\includegraphics[scale=0.32]{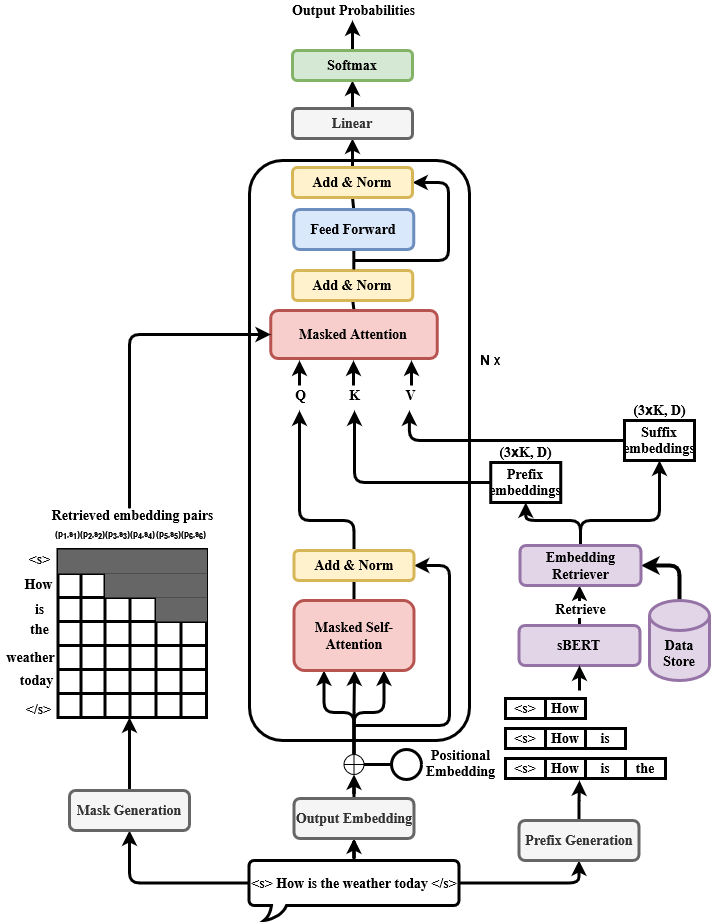}}
  \vspace{-3mm}
\end{minipage}
\caption{SUREALM training with $\delta=1, k=2$.}
\label{fig:proposed}
\vspace{-5mm}
\end{figure}

Another challenge is to make SUREALM fit to mini-batch training where a batch of padded word sequences are fed into the model and different suffix embeddings should be applied at different time $i$. To enable mini-batch training, we first concatenate all offline-retrieved suffix embeddings. Then we construct a suitable attention mask so that each word position $i$ should only attend to the allowed suffix embeddings in masked cross attention, and the previous word positions as in causal LM in masked self attention.
% commented by Wilson
%In regular causal LM, we compute output probability given previous target labels, i.e. \begin{equation}
%    P(X) = \prod_{i=1}^{T} P(x_i|x_{\leq i-1}).
%\end{equation} 
Denote $\mathcal{C}_k^{\bigoplus} (W_{\le i}) = \bigoplus_{i'=1}^i \bigoplus_{j=1}^k \{(p_{i'}^{(j)}, s_{i'}^{(j)})\}$ as a concatenation of all previously retrieved top-k prefix-suffix embedding pairs. Probability of generating a word sequence $W$ becomes:
%In our case, at decoder timestep $i$, we take into account the accumulated embedding pairs %retrieved from all previous timesteps up until $i$. We define the top-$k$ accumulated embedding %pairs up to timestep $i$ an ordered set as \begin{align}
%    \mathcal{C}_k^+ (X_{\leq i}) &= \bigcup_{j=1}^{\lfloor i / N' \rfloor} %\mathcal{C}_k(X_{\leq jN'}) \\ &= \{(p_1,s_1), (p_2,s_2),\dots,(p_{k\lfloor i / N' \rfloor}, %s_{k\lfloor i / N' \rfloor})\}
%\end{align} where $N'$ is the frequency of retrieval. $N'$ is similar to the idea of $N$ defined in section~\ref{subsec:preprocess}. In fact, our experiments have shown that it is most favorable when the incremental step during index is synchronous with the frequency of retrieval during training and inference, i.e. $N'=N$. Given the above definition, our decoding process becomes
\vspace{-2.5mm}
\begin{equation}\label{p}
    P(W) = \prod_{i=1}^{N} P(w_i| W_{\leq i-1}, \mathcal{C}^{\bigoplus}_k(W_{\leq i-1})). 
\end{equation} 
where $W_{\leq i-1} = w_1w_2...w_{i-1}$. SUREALM employs a transformer architecture which follows the Query-Key-Value inputs defined as follows ($i=1,2,3,\dots,|Q|$, $j=1,2,3,\dots,|J|$):
%Given input sequence $X$ and $\mathcal{C}^{+}_k(X_{\leq T-1})$, we add cross-attention blocks %for which we define key(K), query(Q), value(V), and the explicit attention mask(M) as follows:
\begin{align}
%    K &= \text{Concat}(\text{Embedding}(W), p_1, p_2, \dots, p_{k\lfloor i / N' \rfloor}) \\
    Q &= \text{Embedding}(W) \\
    K &= \text{Concat}(p_1, p_2, \dots, p_J) \\
    V &= \text{Concat}(s_1, s_2, \dots, s_J) \\
    M_{i,j} &= \begin{cases}
    0, \text{ $j \leq k (\lceil i / \delta \rceil- 1)$ }\\
    -\infty, \text{Otherwise.}
    \end{cases}
\end{align}
Here $M\in \mathcal{R}^{|Q|\times|J|}$ is an attention mask which masks future positions in keys and values regarding the retrieved suffix embeddings.
%When input sequence is batched, $M$ can be trivially extended by combining $M$ and a padding %mask $P$ that simply masks positions that are padded: $M_{\text{extended}} = M \texttt{\&} P$.
Finally, we obtain the output attention weights in the masked attention block as follows:
\vspace{-3mm}
\begin{equation}
    \text{Attention}(Q,K,V,M) = \text{Softmax}(\frac{QK^{T} + M}{\sqrt{d_k}})V
\end{equation}
where $d_k$ denotes the embedding dimension of the keys and values.
Figure~\ref{fig:workflow} illustrates SUREALM architecture.
%We leverage the cross-attention mechanism by providing aggregated history \& future context in %$K$ and $V$ so that we can train the model to robustly attend to context information when %beneficial. 
Notice that SUREALM can be initialized with weights from any pre-trained transformer of the BERT or GPT model families. We fine-tune the model using the training text to minimize the cross-entropy loss. 
%we leave everything else unchanged, including self-attention blocks, which means %that we can %fine-tune a causal LM by initializing with any pre-trained self-attention %encoder/decoder %architecture.
\vspace{-1mm}
\subsection{SUREALM Decoding}
\label{subsec:decode}
SUREALM decoding is similar to any decoding algorithm in sequence generation except that suffix embedding retrieval is performed when the prefix is updated during generation. We start from the start symbol as the initial prefix. Then suffix embedding retrieval takes place using the current prefix as a query. The top-k suffix embeddings are added into the
set $\mathcal{C}_k^{\bigoplus} (W_{\le i})$ as extra inputs to the transformer in a progressive manner. The next word is generated from the output word distribution and then appended to the prefix. The generation process is repeated until the end-of-sentence symbol is encountered. We follow the decoding algorithm implementation in the Huggingface transformers library~\cite{wolf-etal-2020-transformers} and augment an embedding retriever in our implementation.
%In causal language modeling, same workflow is typically used for both train and test time %(teacher forcing), which can be concluded in \ref{subsec:training}. During generation, however, %we do not have ground-truth labels, so full-sentence pass together with %$\mathcal{C}^{+}_k(X_{\leq T-1})$ is not possible. Therefore, we design a simple algorithm in %\ref{inf} to progressively obtain $\mathcal{C}^{+}_k(X_{\leq i})$ as generation goes on.

\vspace{-3mm}
\section{Experiments}
\vspace{-2mm}
\label{sec:expt}
In this section, we compare SUREALM of different configurations with baseline LMs for sequence generation. We report word perplexity for all experiments as well as the choice of hyper-parameters.
\vspace{-3mm}
\subsection{Setup}
\label{subsec:data}
We used the dataset from the Dialogue System Technology Challenge 9 (DSTC9)\cite{kimdstc9}. The original dataset were designed for evaluating spoken dialogues that involves accessing an external knowledge base containing a list of question-answer pairs about each named entity, such as ``Can I bring my pet to A and B Guest House?'' and ``No, pets are not allowed at this property.''. For language modeling purpose, we treated each dialogue turn as independent sentence and we only kept unique sentences in our training, validation and test sets. Then each sentence was assigned with an unique sentence ID so that they can be uniquely identified in embedding retriever. Our resulting training dataset contained 126,877 unique dialogue turns mentioning 145 named entities covering four domains: hotel, restaurant, taxi and train. Our validation dataset contained 18,037 unique dialogue turns.
The test dataset had 18,021 unique dialogue turns covering 523 unseen named entities including a new domain on attraction. Due to the introduction of a new domain, we further split the test dataset into in-domain and out-of-domain portions and only evaluated on the in-domain portion. Since the test turns did not have the manual domain label, we applied named entity recognition on all dialogue turns and we used the detected named entity and applied its corresponding domain label to dialogue turns. The question-answering knowledge base was not added into our data store. The data store only contained the prefix-suffix embeddings from the training sentences.

For data preprocessing, we followed Section~\ref{subsec:preprocess} to generate prefix-suffix pairs of the training dataset, pre-compute the embedding pairs with pre-trained sentence transformers~\cite{reimers-2019-sentence-bert}, and index and store them using FAISS~\cite{johnson2019billion}. We also pre-computed the prefix embeddings for validation/test sets to accelerate the retrieval process. To avoid cheating, such embeddings were not indexed in FAISS.

\vspace{-4mm}
\subsection{Training details}
%\subsubsection{Setting}
In SUREALM, there are two modeling components to consider: (1) encoding model; (2) language model. For the encoding model, we used pre-trained sentence transformers~\cite{reimers-2019-sentence-bert} to encode prefixes and suffixes. We tried small-scale models~\footnote{384-dimension with 6 to 12 layers} and standard-scale models~\footnote{768-dimension with 12 layers} respectively in our experiments. For the language model, we employed transformer-based language models with various weight initialization strategies. Inspired by~\cite{rothe-etal-2020-leveraging}, we explored different sentence transformer checkpoints to initialize the language model weights.
%We used Sentence-BERT pre-trained models hosted on Huggingface. We trained SUREALM on both small-scale (6-layer, 384-dimension) and standard-scale (12-layer, 768-dimension) transformer architectures.
On small-scale model training, we used a batch size of 128, \texttt{AdamW} optimizer with learning rate of \texttt{2e-5}, and linear learning rate scheduler with 500 warmup steps. On standard-scale model training, we used a batch size of 64 and learning rate of \texttt{1e-5} and kept the same settings as in the small-scale model training. Since our dataset was relatively small, we trained SUREALM for 200 epochs and chose the model with the minimum validation perplexity.

In our preliminary experiments, we chose the best hyper-parameters based on the validation perplexity. Results showed that it was crucial to retrieve at each prediction time step ($\delta=1$). We chose $m=10$ for suffix truncation and $K=8$ for retrieving the top-k prefix-suffix embeddings, yielding the best perplexity results. We fixed these hyper-parameters for further experiments below.

% Then we investigated the impact of weight initialization for both the sentence transformer and the LM. We also tried using more than one datastore each constructed with different sentence transformers. We did the experiments above on small-scaled models.  

%We finally tested the scalability of our method by applying hyper-parameters from small-scaled experiments on various standard-scaled architectures for which . We also evaluated the resulting models on test data by plugging in datastore obtained from test knowledge base as in \ref{test}.

%\subsubsection{Metric \& baselines}
%We used \textbf{perplexity} on target labels as our metric. As most of our experiments focused on fine-tuning established language models, we directly compared the performance of those models fine-tuned with and without our method.
\vspace{-2mm}
\subsection{Results}

\subsubsection{Small-scale models}
For baselines, we fine-tuned transformer-based masked LMs (Mini-LM~\cite{10.5555/3495724.3496209}) of 6 and 12 layers initialized with pre-trained weights. For encoding, we used pre-trained sentence transformers (\texttt{multi-} \texttt{qa-MiniLM} and \texttt{all-MiniLM} )\footnote{In Table~\ref{table:small}, \texttt{Multi} stands for the former, and \texttt{All} stands for the latter.}. Our LMs were initialized with the weights of the pre-trained sentence transformers or the masked LMs.
%We initialized LMs with weights from the corresponding ST, MiniLM, or no weight initialization at all. 
Results in Table~\ref{table:small} show that:
\begin{enumerate}
    \item SUREALM achieved lower perplexity compared to baselines in all experiments. Our best small model achieved relative test perplexity reduction of 19\% compared to the baseline.
    \item Note that we compared MiniLM based SUREALM with that initialized with Bert as they shared the same vocabulary. While SUREALM initialized with BERT achieved the best performance (21\%), our small-scale models still achieved similar performance gain with 63\% less parameters.
    \item LM weights can be initialized differently from the pretrained ST used for encoding without any performance degradation. This implies flexibility for different weight combinations.
    
%    using a different  SUREALM can achieve the same or better performance when initialized with weights of a general-purpose LM instead of its corresponding ST, which potentially allows us to fine-tune any general-purpose LM not limited to the BERT family, including autoregressive LMs like GPT.
%    \item SUREALM also achieves faster convergence in most of our experiments.
\end{enumerate}

\vspace{-4mm}
\subsubsection{Standard-scale models}
We then compared standard-scale SUREALM with popular state-of-the-art LM baselines(BERT, RoBERTa and GPT2) of which the weights were used for initialization. However, since they used different tokenizers resulting in different output vocabulary sizes, we could only compare models with  with similar vocabulary sizes. Bert initialized SUREALM in Table~\ref{table:small} achieved best relative test perplexity reduction by 21\% across all experiments.
Table~\ref{table:big} shows perplexity results with increased vocabulary size of 50k. SUREALM achieved relative test perplexity reduction by 13\% and 19\% compared to GPT2 and RoBERTa baseline respectively.We also experimented with the distilled version of the two architectures yielding consistent results. Note that some experiments with GPT2, while not bringing significant perplexity reduction, achieved matching performance with the baseline. %maybe explain this in discussion?

%We employed \texttt{multi-qa-distilbert-cos-v1}, \texttt{all-distilroberta-v1} and \texttt{multi-qa-mpnet-base-cos-v1} as our STs, and we initialized SUREALM with BERT family LMs such as RoBERTa, BERT, as well as GPT-2. We report our experiment results in Table~\ref{table:big}. SUREALM still outperformed all baseline LMs under all weight initialization. However, the choice of ST and LM weights affected performance significantly. For example, \texttt{multi-qa-distilbert-cos-v1} was the best at providing useful embedding pairs for all 3 LMs. SUREALM with BERT-base improved its baseline by 0.54 in test perplexity. While still outperforming their baselines, our experiments on RoBERTa and GPT-2 did not yield improvement margin as great as that of BERT or small-scale models. Nevertheless, the consistent improvement compared with baselines shows that, given a pre-trained sentence transformer, SUREALM is capable of training an causal LM from mass pre-trained LMs with performance improvement even if the sentence transformer has different tokenization, vocabulary, and weights than the LM. We also argue that the choice of model weights is crucial and more investigation is needed. 

\begin{table}[htb] 
%\label{table:small}
\vspace{-3mm}
\centering
\caption{Validation and test perplexities on 30K vocabulary. ST stands for sentence transformers. LM init. stands for weight initialization for our LMs. RSR gives the relative size ratio of our LMs compared with that of BERT-base.}
%\resizebox{\textwidth}{15mm}{oik
\begin{tabular}{|c|c|c|c|c|}
\hline
Pretrained ST     & LM init.     & Val. ppl    & Test ppl    & RSR    \\
\hline
%(Baseline)           & No initialization      & 5.78   & 8.94     & 200  \\
NA (baseline)          & All-L6        & 5.36 & 8.04 & 0.21\\
NA (baseline)          & Multi-L6    & 5.35  &7.97 & 0.21\\
NA (baseline)         & MiniLM-L6     & 5.43   & 7.94 & 0.21\\ %    & 100
NA (baseline)          & All-L12       & 5.33 & 7.96 & 0.30\\
NA (baseline)         & MiniLM-L12     & 5.31  & 7.84 & 0.30\\
NA (baseline)          & BERT-base     & 5.19 & 7.58 & 1.0\\
\hline
%Multiqa-MiniLM            & No initialization     & 5.32 &  8.17    & 72   \\ 
%All-MiniLM & No initialization & 5.34 & 8.10 & 63\\
Multi-L6           & Multi-L6      & 4.54   & 6.81  & 0.24\\   %& 63   \\ 
Multi-L6            & MiniLM-L6      & 4.71   & 6.85   & 0.24  \\ % & 62   \\ 
All-L6          & All-L6      & 4.94   &   7.19 & 0.24\\ %  & 145  \\
All-L6          & MiniLM-L6      & 4.85   &   7.03  & 0.24\\ % &  \text{131}\\
All-L6          & All-L12            & 4.46  & 6.67 & 0.37\\
All-L6            & MiniLM-L12      & 4.48  & 6.62 & 0.37\\
Multi-L6          & All-L12       & 4.32  & 6.53 & 0.37\\
Multi-L6        & MiniLM-L12      & 4.29  & 6.47 & 0.37\\
All-L12             & All-L12       & 4.24  & 6.42 & 0.37\\
All-L12             & MiniLM-L12    & 4.18  & 6.37 & 0.37\\
Multi-distilbert     & BERT-base    & \textbf{4.07}  &\textbf{6.02} & 1.0\\
\hline
\end{tabular}%}
\label{table:small}
\vspace{-3mm}
\end{table}
\begin{table}[htb!] 
%\label{table:big}
\vspace{-3mm}
\centering
\caption{Validation and test perplexities on 50K vocabulary. ST stands for sentence transformers. LM init. stands for weight initialization for our LMs. RSR gives the relative size ratio of our LMs compared with that of GPT2-base.}
%\resizebox{\textwidth}{15mm}{
\begin{tabular}{|c|c|c|c|c|}
\hline
Pretrained ST     & LM init.     & Val. ppl    & Test ppl  & RSR    \\ %   & Epochs \\
\hline
N/A (baseline)           & distilroberta    & 5.47  & 7.88 & 0.54\\
N/A (baseline)          & distilgpt2        & 5.18  & 7.72 & 0.54\\
N/A (baseline)           & RoBERTa-base      & 5.35   & 7.69 & 1.0\\ %     & 42  \\
N/A (baseline)           & GPT2-base     & 5.01   & 7.60 & 1.0  \\ %  & 71  \\
\hline

All-distilroberta           & distilroberta    & 4.56  & 6.49 &  0.63\\
Multi-distilbert          & distilgpt2        & 5.12  & 7.77 & 0.63\\
All-distilroberta           & RoBERTa-base      & \textbf{4.32}   &   \textbf{6.26} & 1.23 \\ % & 26   \\ 
Multi-distilbert           & RoBERTa-base      & 4.47   &   6.52 &1.23 \\ % & 26   \\ 
Multi-mpnet          & GPT2-base      & 5.33   &  7.54  & 1.23\\ %    & 22  \\
Multi-distilbert         & GPT2-base     & 4.76   & 6.63 & 1.23 \\ % & 22 \\
All-distilroberta         & GPT2-base       &  5.36  &  7.66 & 1.23 \\ %   & 22 \\
%Multiqa-distilbert            & BERT-base     & \textbf{4.88} &  \textbf{7.04}    & \textbf{18}  \\ 

\hline
\end{tabular}%}
\label{table:big}
\vspace{-6mm}

\end{table}
% Add empirical sequence generation results using some prefix

\vspace{-3mm}
\section{Discussions}
\vspace{-2mm}
\label{sec:discuss}
% talk about the observation of excluding the current word.
During embedding retrieval, we investigated the inclusion of the current word into the suffix in a training sentence, meaning that we only split a training sentence into prefix and suffix instead of prefix, current word and suffix mentioned in Section~\ref{subsec:preprocess}. Then we followed the same procedure to encode the prefixes and suffixes and reran SUREALM training and evaluation. However, we observed no test perplexity reduction compared to the baseline. Excluding the current word from suffix may be analogous to applying a mask token in the mask LM. After excluding the current word, SUREALM focuses on information from the word history and the retrieved suffix context for word prediction. It is possible that the embedding retrieval results may contain sentences that share similar prefixes but having an identical suffix as in the current input sentence. From this perspective, excluding the current word from suffix is reasonable to avoid SUREALM from overly relying on the suffix embeddings and forgetting the word history in word prediction.

%\subsubsection{Domain adaptation from new knowledge base}
%Without additional fine-tuning, we also evaluated on out-of-domain test set with our best small-scale SUREALM by allowing it to retrieve from data store created using test knowledge base. In our experiment, we retrieved 4 top-K embeddings from each data store (train and test). However, we did not see significant improvement over baseline. The out-of-domain test set we have extracted from the original test dataset has only 3709 utterances and contains simulated speech disfluencies such as "ok ummm how much is the cost of park at the courtyard". So we suspect that evaluation on this test set can be inconclusive and SUREALM still has the potential to be able to adapt to new domains by introducing new knowledge bases.

%\subsubsection{Better control for retrieval process}
%While we have employed some tricks to ensure that our retrieval function could return useful prefix-suffix pairs for our LM, there are still potential issues. For example, at timestep $i$, SUREALM does not check if the embedding pairs retrieved at the current timestep have appeared in previous steps, i.e.  $\mathcal{C}_K^{\bigoplus} (W_{\le i}) = \bigoplus_{i'=1}^i \bigoplus_{k=1}^K \{(p_{i'}^{(k)}, s_{i'}^{(k)})\}$ is not a set. This might bring redundent information hindering SUREALM from generalizing. In addition, our current implementation using FAISS makes it so that embeddings in data store with the same cosine similarity score will be retrieved arbitrarily instead uniformly so it is possible that  In future work we will also address 
\vspace{-3mm}
\section{Conclusions}
\vspace{-2mm}

We have proposed a suffix retrieval-augmented language model to simulate bi-directional contextual effect while remains autoregressive so that our model can be used for sequence generation. Our proposed model shows promising perplexity performance compared to state-of-the-art LM baselines. In the future, we plan to evaluate our model on large corpora. In addition, we plan to extend our model on conditional generation such as dialogue response generation. Lastly, we will investigate domain LM adaptation using our proposed model.

%\section{FOOTNOTES}
%\label{sec:foot}

%Use footnotes sparingly (or not at all!) and place them at the bottom of the
%column on the page on which they are referenced. Use Times 9-point type,
%single-spaced. To help your readers, avoid using footnotes altogether and
%include necessary peripheral observations in the text (within parentheses, if
%you prefer, as in this sentence).

% Below is an example of how to insert images. Delete the ``\vspace'' line,
% uncomment the preceding line ``\centerline...'' and replace ``imageX.ps''
% with a suitable PostScript file name.
% -------------------------------------------------------------------------
%\begin{figure}[htb]
%
%\begin{minipage}[b]{1.0\linewidth}
%  \centering
%  \centerline{\includegraphics[width=8.5cm]{image1}}
%%  \vspace{2.0cm}
%  \centerline{(a) Result 1}\medskip
%\end{minipage}
%%
%\begin{minipage}[b]{.48\linewidth}
%  \centering
%  \centerline{\includegraphics[width=4.0cm]{image3}}
%%  \vspace{1.5cm}
%  \centerline{(b) Results 3}\medskip
%\end{minipage}
%\hfill
%%\begin{minipage}[b]{0.48\linewidth}
% \centering
%  \centerline{\includegraphics[width=4.0cm]{image4}}
%%  \vspace{1.5cm}
%  \centerline{(c) Result 4}\medskip
%\end{minipage}
%%
%\caption{Example of placing a figure with experimental results.}
%\label{fig:res}
%%
%\end{figure}

%\vfill
%\pagebreak

\vfill\pagebreak

\bibliographystyle{IEEEbib}
\bibliography{strings,refs}

\end{document}